# Evaluating Open-Source Vision Language Models for Facial Emotion Recognition against Traditional Deep Learning Models

Vamsi Krishna Mulukutla[1], Sai Supriya Pavarala[1], Srinivasa Raju Rudraraju[1] and Sridevi Bonthu[1,*]

[1]Department of Computer Science and Engineering, Vishnu Institute of Technology, Bhimavaram

## Abstract

Facial Emotion Recognition (FER) is crucial for applications such as human-computer interaction and mental health diagnostics. This study presents the first empirical comparison of open-source Vision-Language Models (VLMs), including Phi-3.5 Vision and CLIP, against traditional deep learning models—VGG19, ResNet-50, and EfficientNet-B0—on the challenging FER-2013 dataset, which contains 35,887 low-resolution, grayscale images across seven emotion classes. To address the mismatch between VLM training assumptions and the noisy nature of FER data, we introduce a novel pipeline that integrates GFPGAN-based image restoration with FER evaluation. Results show that traditional models, particularly EfficientNet-B0 (86.44%) and ResNet-50 (85.72%), significantly outperform VLMs like CLIP (64.07%) and Phi-3.5 Vision (51.66%), highlighting the limitations of VLMs in low-quality visual tasks. In addition to performance evaluation using precision, recall, F1-score, and accuracy, we provide a detailed computational cost analysis covering preprocessing, training, inference, and evaluation phases, offering practical insights for deployment. This work underscores the need for adapting VLMs to noisy environments and provides a reproducible benchmark for future research in emotion recognition.

## 1. Introduction

Facial Emotion Recognition (FER) has emerged as a critical area in artificial intelligence, with applications spanning human-computer interaction, behavioral analysis, and surveillance systems [1]. Traditional deep learning models such as VGG19, ResNet-50, and EfficientNet-B0 have demonstrated strong performance in FER due to their ability to extract robust visual features [2]. However, the widely used FER-2013 dataset presents unique challenges, including low resolution, class imbalance, and varying lighting conditions, which complicate model evaluation [3].

While deep learning models have been successful under such conditions, recent advances in Vision-Language Models (VLMs)—such as Phi-3.5 Vision, LLaMA-3.2 Vision Instruct, and CLIP—have raised interest in their potential to enhance FER through multi-modal understanding and large-scale pretraining [4]. Despite their success in general vision tasks, VLMs remain underexplored for FER, particularly in noisy and low-resolution environments. Moreover, although face restoration techniques have been studied independently, there is limited research on combining them with VLM-based FER pipelines.

This study addresses that gap by introducing a novel pipeline that integrates GFPGAN-based image restoration with FER evaluation using open-source VLMs. We empirically compare their performance with traditional deep learning models on the FER-2013 dataset, revealing that the latter still outperform VLMs—likely due to VLMs' reliance on structured, high-quality data that struggles with real-world visual variability. In addition to performance benchmarking using accuracy, precision, recall, and F1-score, we also analyze the computational cost across preprocessing, training, and inference stages to assess real-world deployment viability.

By enhancing FER-2013 images through GFPGAN and evaluating multiple architectures, this work provides new insights into the limitations of current VLMs in FER tasks and offers a reproducible benchmark for future studies aiming to adapt large models to challenging vision applications. The key contributions of this work are as follows:

- Systematically evaluated traditional deep learning models (VGG19, ResNet-50, EfficientNet-B0) and VLM-based vision models (Phi-3.5 Vision, LLaMA-3.2 Vision Instruct, CLIP-ViT-B/32) on the FER-2013 dataset, offering a comprehensive performance comparison.
- Provided empirical findings that highlight the challenges VLMs face in recognizing facial emotions from low-resolution and heterogeneous images.
- Identified data resolution and structure as critical factors influencing model generalization, showing that VLMs trained on clean datasets struggle with real-world FER conditions.
- Introduced a novel GFPGAN-enhanced preprocessing pipeline and conducted a comparative computational cost analysis, offering practical insights into the trade-offs between accuracy and efficiency for real-world deployment.

## 2. Literature Review

FER has been extensively studied using Convolutional Neural Networks (CNNs), which have achieved considerable success in classifying facial expressions from images [8, 9]. For example, Jaiswal et al. [10] demonstrated that deep CNNs can achieve validation accuracies of 70.14% on FER-2013 and 98.65% on JAFFE, emphasizing the role of data quality in model performance. The foundational work of Krizhevsky and Hinton [11] on deep CNNs, although initially trained on CIFAR-10, laid the groundwork for modern FER models through innovations in feature extraction and visualization of learned filters.

Researchers have also explored FER applications in other domains. Bartlett and Movellan [12] applied facial analysis for drowsiness detection in drivers, leveraging facial action units to monitor fatigue, which highlights the safety-critical potential of FER systems. To improve performance further, hybrid architectures have been proposed. Al-Shabi et al. [13], for instance, combined CNNs with SIFT features and applied extensive data augmentation and model aggregation techniques, achieving strong performance on FER-2013 and CK+ datasets.

Recent works have focused on FER under real-world conditions such as occlusion, lighting variation, and low resolution. Image restoration techniques like GFPGAN [7] have been successfully employed to enhance low-quality facial images, thereby improving downstream classification accuracy. However, few studies have extended such preprocessing techniques to VLMs. While models like CLIP and Flamingo [14] have shown promise in general emotion recognition tasks, their performance degrades when applied to noisy, unstructured datasets. Other methods have explored adversarial learning and domain adaptation to improve generalization [15, 16].

Despite these advancements, there remains a significant gap in evaluating VLMs under degraded image conditions typical of real-world FER datasets. Specifically, little research has investigated how restoration pipelines like GFPGAN can be combined with VLMs to improve robustness. This study addresses this gap by integrating GFPGAN-based preprocessing with state-of-the-art VLMs (e.g., Phi-3.5 Vision and CLIP) and benchmarking them against traditional deep learning models on the FER-2013 dataset. Our work provides new insights into the performance trade-offs between these model families under constrained visual conditions.

## 3. Requirements

To conduct the experiments in this study, appropriate computational resources and software environments were essential. The hardware setup included systems with at least an Intel Core i5 processor or equivalent, although GPU acceleration was strongly recommended for training and inference tasks. For general experimentation, NVIDIA T4 GPUs were used via cloud-based platforms such as Google Colab. For more computationally intensive models and preprocessing operations (e.g., GFPGAN), higher-end GPUs such as NVIDIA A100 or V100 were utilized through Colab Pro.

Deep learning tasks required a minimum of 12GB RAM, with 25GB or more preferred for optimal performance and stability during training. Cloud storage solutions, such as Google Drive, were used for efficient dataset and model management.

The software environment was based on Python (version 3.7 or higher), and included essential libraries such as Hugging Face Transformers (for pre-trained VLM integration and inference) and Pillow (PIL) for image processing. All model evaluations—including Phi-3.5 Vision and CLIP—were conducted using publicly available implementations hosted in open-source repositories. Google Colab provided the necessary compute infrastructure for executing and scaling these experiments.

A summary of compute time, GPU usage, and memory demands across models and tasks is presented in Section 6.3, to support reproducibility and practical deployment planning.

## 4. Datasets

This study uses the FER-2013 dataset, a widely recognized benchmark for facial emotion recognition. It comprises 35,887 grayscale facial images, each with a resolution of 48 × 48 pixels. These images are categorized into seven emotion classes: angry, disgust, sad, happy, neutral, surprise, and fear [23]. Collected from diverse sources, the dataset includes variations in facial expressions, poses, occlusions, and lighting conditions—factors that make it particularly challenging for automated emotion recognition.

FER-2013 was chosen for its ability to simulate real-world scenarios through low-quality images, significant class imbalance, and noisy visual content. These characteristics provide a rigorous testbed for evaluating both traditional deep learning models (e.g., ResNet-50, EfficientNet-B0) and Vision-Language Models (e.g., Phi-3.5 Vision, CLIP). By assessing these models under the same challenging conditions, we aim to understand their respective capabilities in handling imprecise, low-resolution facial data.

The dataset is divided into training, validation, and test sets, ensuring a fair and standardized comparison across models. Due to the noisy and imbalanced nature of the dataset, we applied preprocessing techniques such as grayscale scaling, contrast enhancement, and noise filtering to improve feature extraction. Figure 1 shows representative sample images from the FER-2013 dataset.

Overall, FER-2013 provides a comprehensive and practical benchmark for evaluating model robustness, particularly when adapting large-scale pretrained models like VLMs to specialized vision tasks such as FER in unconstrained environments.

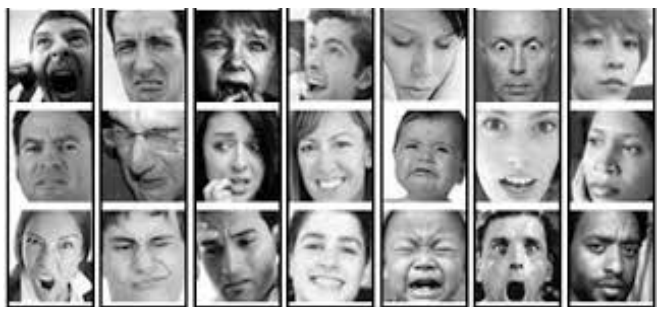

**Figure 1**. Sample images from the FER-2013 dataset, illustrating noise, resolution challenges, and emotion variability.

## 5. Methodology

The methodology adopted for this study is designed to explore the impact of preprocessing techniques, particularly GFPGAN, on the performance of traditional deep learning and VLMs in FER using the FER-2013 dataset. The following subsections describe the stages involved in preprocessing, model selection, training, evaluation, and computational cost analysis.

### 5.1 Pre-processing

The FER-2013 dataset comprises low-resolution grayscale facial images, many of which suffer from issues like blur, occlusions, poor contrast, and inconsistent lighting. These challenges impair the models' ability to extract meaningful features, reducing classification performance. To address these limitations, a robust preprocessing pipeline was established, involving both image enhancement and data filtering.

The cornerstone of the preprocessing pipeline is GFPGAN, a deep learning-based facial image restoration model. GFPGAN is designed to recover lost facial details, rectify distortions, and improve the overall clarity of facial images while maintaining identity preservation. Its architecture integrates a coarse-to-fine facial reconstruction network with a pre-trained face prior module, enabling effective restoration of degraded facial features. Figure 2 illustrates the complete methodology employed by GFPGAN.

By leveraging GFPGAN, the dataset's low-quality images were enhanced, allowing the models to detect finer facial details, thereby improving the learning process and emotion classification accuracy. Alongside enhancement, a data filtering step was introduced to remove images with severe distortions, occlusions (e.g., faces obscured by hair), or blank entries caused by data acquisition issues. Figure 3 demonstrates the transformation of low-resolution grayscale images to restored RGB outputs. A combination of automatic quality-checking tools and manual inspection was used to ensure only high-quality, relevant images remained.

This preprocessing phase significantly improved the quality and consistency of the dataset, thereby enhancing the robustness and accuracy of both deep learning and VLM models. The enriched dataset enabled the models to learn more representative features and make reliable emotion predictions under real-world conditions.

### 5.2. Model Selection

To comprehensively evaluate the effectiveness of traditional deep learning models and Vision-Language Models in FER, five distinct models were selected: VGG19, ResNet-50, EfficientNet-B0, Phi-3.5 Vision, and CLIP. These models were chosen based on their architectural diversity, performance history, and capability to handle the unique challenges posed by the FER-2013 dataset.

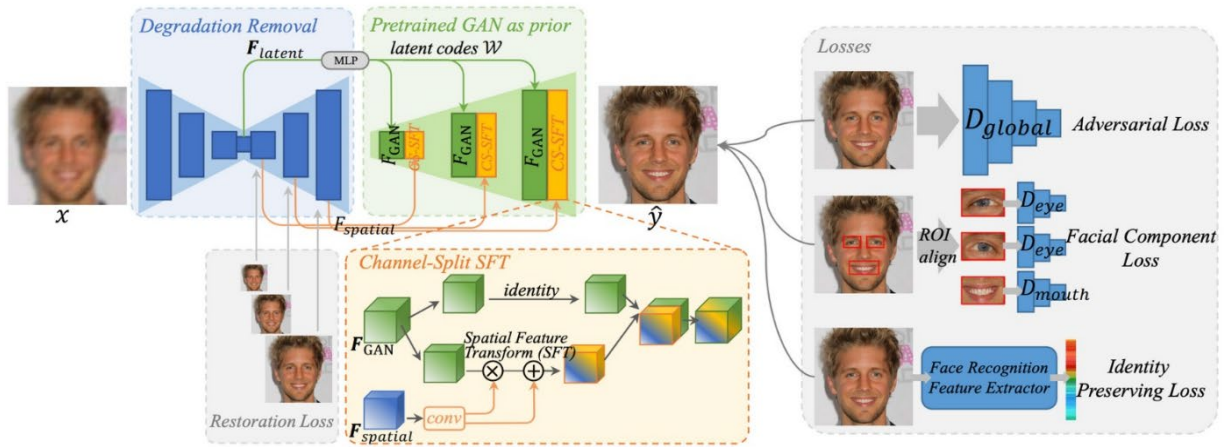

**Figure 2.** The Methodology of GFPGAN [7].

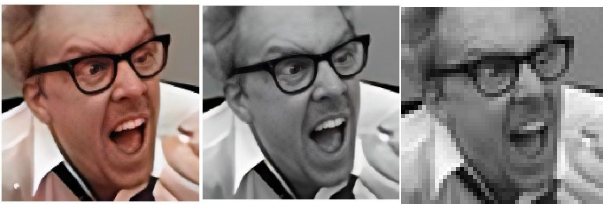

**Figure 3.** This figure demonstrates the low to high resolution image, and followed by Grey to RGB image.

## 5.2. Model Selection

To comprehensively evaluate the effectiveness of traditional deep learning models and Vision-Language Models in FER, five distinct models were selected: VGG19, ResNet-50, EfficientNet-B0, Phi-3.5 Vision, and CLIP. These models were chosen based on their architectural diversity, performance history, and capability to handle the unique challenges posed by the FER-2013 dataset.

### 5.2.1. VGG19

VGG19 stands out as a deep CNN with a straightforward yet powerful design. It's built with 19 layers and leans heavily on compact 3×3 convolutional filters, which excel at picking up subtle image details [26]. This setup has proven its worth in image classification, and it's particularly handy for facial emotion recognition, thanks to its knack for pulling out everything from basic edges and textures to complex facial features and expressions [27]. Sure, it's a bit of a resource hog compared to sleeker modern models, but VGG19 holds its own as a reliable benchmark, especially when tackling the diverse facial expression shifts in the FER-2013 dataset with its robust feature extraction skills.

### 5.2.2 ResNet-50

ResNet-50 (a 50-Layered Residual Network)[28] is one of the commonly applied deep learning models that uses residual learning for overcoming the vanishing gradient problem of deep networks. ResNet-50 utilizes skip connections (shortcuts) to enable gradients to pass through more easily in backpropagation, hence enabling deep networks to learn more conveniently [29]. ResNet-50 is well known for its ability to extract strong features and has also shown very high accuracy on other image classification problems, making it a strong contender for emotion recognition in faces. Due to its deep model structure, intricate features such as edges and textures and prominent features such as face structures and expressions of the FER 2013 dataset can be extracted.

### 5.2.3. EfficientNet-B0

EfficientNet-B0 belongs to the EfficientNet model family, which is computationally highly efficient and accurate with low computation [30]. It proposes a compound scaling approach in which the model scales depth (layer wise), width (channels per layer), and resolution (image size) simultaneously for performance. EfficientNet-B0, in comparison to regular CNNs, provides higher accuracy with fewer parameters, which is beneficial in situations where computational resources are scarce. In FER 2013, EfficientNet-B0 was chosen since it can learn fine-grained features with low weight and speed.

### 5.2.4. Phi-3.5 Vision

A large multimodal model that can reason and comprehend images [31]. Unlike normal CNNs, which are only concerned with pixel-level feature extraction, Phi-3.5 Vision uses pretrained vision knowledge from huge vision datasets such that it can recognize emotions more contextually. It is better than traditional models that utilize transfer learning and fine-tuned embeddings with noisy, low-quality images. This therefore makes it a powerful tool for FER 2013, especially where common deep models are not comfortable with ambiguous or unclear expressions.

### 5.2.5. CLIP (Contrastive Language-Image Pretraining)

CLIP is a refined vision-language model from OpenAI pre-trained from natural language supervision rather than labeled data [32]. It is trained on ample text-image pairs so that it can generalize across various vision-related tasks, including emotion recognition. In contrast to CNNs that are primarily based on labeled marks, CLIP can map textual explanations to images and thus is very flexible in detecting facial expressions even with limited training data. Its capacity to identify emotions from a general point of view offers richer insights compared to deep CNN-based models [33].

## 5.3. Training

### 5.3.1. Deep Learning Models

**VGG19:** VGG19 was trained for 60 epochs, taking advantage of its deep convolutional layers to extract layer-wise features from the images. The model employed ReLU activation function in the convolutional layers to bring non-linearity and enhance feature representation. Batch normalization was applied after every convolutional layer to promote stable training and quicker convergence. Dropout was added as a regularization method to avoid overfitting by randomly dropping out neurons throughout training. The weights of the model were initialized with He-uniform kernel initialization to better distribute the weights. The last classification layer utilized the softmax activation function to provide probability distributions over the seven emotion classes, and categorical cross-entropy was used as the loss function. The Adam optimizer was put to use to update the model's parameters efficiently and improve learning.

**ResNet-50:** ResNet-50 was trained for 60 epochs, utilizing its deep residual connections to strengthen feature learning while preventing vanishing gradient problems. The model employed ReLU activation in the hidden layers to inject non-linearity, helping with improved representation learning. For stable training and quick convergence, batch normalization was implemented following convolutional layers. Dropout was also included as a regularizing method to avoid overfitting by randomly disengaging neurons throughout training. The weights of the model were initialized with He-uniform kernel initialization, which assists in effective weight allocation. The last classification layer utilized the softmax activation function to create probability distributions for the final emotion classes, while categorical cross-entropy was utilized as the loss function. The Adam optimizer was added to effectively update the model's parameters.

**EfficientNet-B0:** EfficientNet-B0 was trained for 30 epochs, taking advantage of its compound scaling method to optimize depth and width for better computational efficiency. Like ResNet-50, it used ReLU activation, batch normalization, dropout, and He-uniform kernel initialization for amelioration of training performance. The last classification layer also used softmax activation, with categorical cross-entropy for the loss function and Adam optimize to update the model's weights efficiently.

### 5.3.2 Vision-based Language Models (VLMs)

Pre-trained vision-based language models like Phi-3.5 Vision and CLIP were also tested for facial emotion detection [34]. These models, however, are already pre-trained on massive multimodal datasets and do not require further training or fine-tuning [35]. Rather, they can be used directly to the dataset, with their zero-shot and few-shot learning ability being tested for comparison of performance with the deep learning models [36]. Recent advancements in Large Language Models are becoming more trustworthy and responsible [37].

**Phi-3.5 Vision Model:** The Microsoft's Phi-3.5 vision-instruct model, crafted by Microsoft, marks a leap forward in compact multimodal AI, blending text and image processing within a lean 4.2-billion-parameter framework. Unlike its siblings—Phi-3.5 mini-instruct and Phi-3.5 MoE-instruct—these variant shines with a 128,000-token context window and excels in tasks like image comprehension, OCR, and multi-frame analysis. Built from scratch with synthetic data and curated public sources, it is fine-tuned through supervised learning and preference optimization for precision and safety.

Performance-wise, Phi-3.5-vision-instruct punches above its weight, scoring 57.0 overall in benchmarks—outpacing peers like LlaVA-Interleave-Qwen-7B (53.1) in forensic detection (92.4) and art style recognition (87.2) [38]. Yet, it stumbles in abstract reasoning, with scores like 29.2 in functional correspondence lagging behind heavyweights like GPT-4o.

**CLIP:** CLIP is an OpenAI-developed multimodal artificial intelligence model that achieves vision-language parity. CLIP is trained on a vast dataset of text-image pairs so that it understands images and text and relates the two modalities. CLIP does this by learning a common latent space representation for images and text. This enables it to accomplish a multitude of tasks, from zero-shot image classification, object detection, and even text-based search for images, without task-specific training [36].

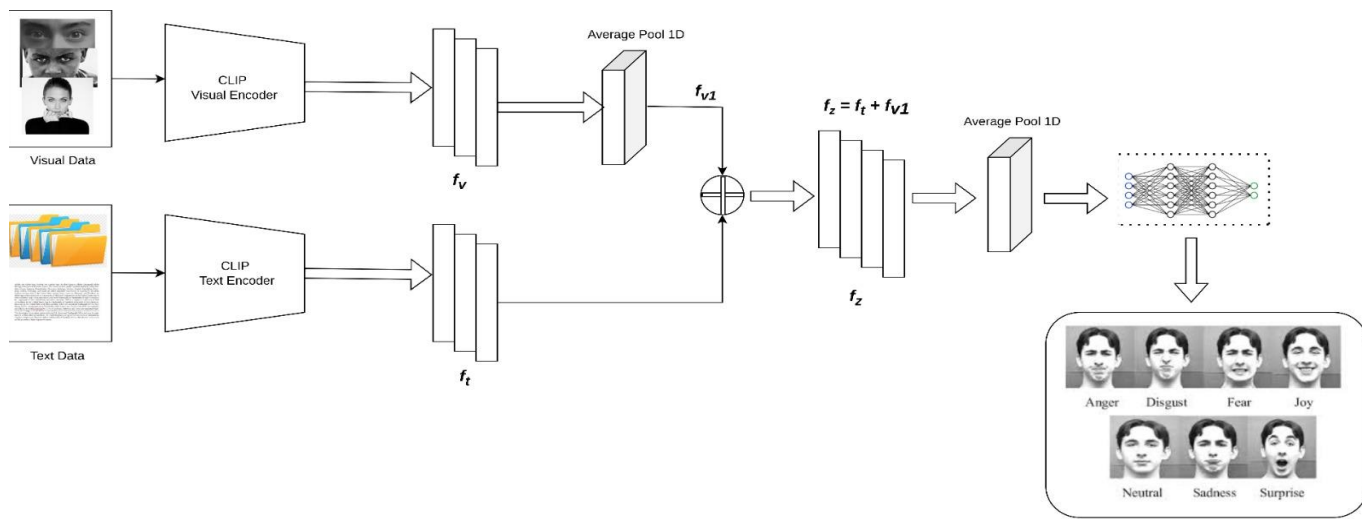

**Figure 4.** The methodology of the CLIP architecture [14].

Architecturally, as shown in Fig. 4, CLIP employs distinct neural networks for processing images and text. The image encoder typically is a vision transformer (ViT) or a ResNet model, while the text encoder is a transformer-based architecture such as GPT models. Through learning over a broad and diverse collection of image-text pairs, CLIP can excel at learning general visual concepts from text descriptions and thus is extremely flexible across tasks without task-specific fine-tuning on the dataset.

From a parameter standpoint, CLIP's base model holds around 400 million parameters, which is proportionally small to newer versions like Phi-3.5 Vision, with billions of parameters. Though small, CLIP performs incredibly well, particularly in the likes of applications such as zero-shot classification, where it is able to categorize images by classes under text prompts without the need for specialized prior training. The performance of CLIP can differ depending upon the task, and it performs optimally at locations where image and text data have a high correspondence.

CLIP has provided good performance [39] on the majority of benchmarks, especially for tasks involving matching an image with its matching text description. The most wonderful thing about CLIP is that it generalizes so well across domains, i.e., it can process an enormous variety of visual concepts without being trained for them in particular. But CLIP is not yet in the state of being able to solve abstract thinking or higher-order multi-step thinking problems compared to other models such as GPT-4 or Phi-3.5 Vision, which are solely designed for carrying out such processes. CLIP is nonetheless still a very formidable tool for a titanic scope of multimodal AI applications with respect to those restrictions.

## 5.4. Evaluation

To measure the models' performance, we employed four main evaluation parameters: precision, recall, F1-score, and accuracy. These parameters provide a clear evaluation of an algorithm's performance, particularly with respect to tackling the class imbalance of the FER2013 dataset.

*Accuracy* counts the total accuracy of an algorithm by determining the total number of correctly categorized classes to the total number of classes. Even though accuracy is a most basic measure, it may not show the actual performance of the model because a dataset may contain imbalances. *Precision* calculates the number of true positive samples to the total of samples predicted positive. It has application when there should be less number of false positives such that neutral or weak expressions are not mistakenly recognized as emotions such as anger or fear by the model. *Recall* is how well a model can classify every possible case of a given class. Higher recall essentially means that an algorithm is adequate at detecting most cases of a particular emotion, which is very much a necessity when there are fewer training samples for a particular emotion like "disgust." *F1-score* is a balanced measurement between recall and precision, which is particularly helpful when class imbalances need to be handled. It takes both false negatives and false positives, providing a better complete understanding of the model performance.

The effectiveness of regular deep learning-based models, such as ResNet-50 and Efficientnet-B0, is compared against the capabilities of vision language models, such as Phi-3.5 Vision and CLIP, in determining facial expression recognition workloads through comparison of such metrics [40, 41].

## 5.5 Computational Cost Overview

In order to assess the resource requirements of various models used in our experiments, we tracked the GPU time, batch sizes, number of epochs, and memory utilization during preprocessing, training, inference, and evaluation stages. This helps establish a compute-efficiency profile for each model and provides deployment-relevant insights.

## 6. Inference, Results and Discussion

We evaluated the performance of three deep learning architectures—VGG19, ResNet50, and EfficientNet-B0—on the FER2013 facial expression recognition dataset, focusing on their ability to classify input images into seven categories of emotion i.e., anger, fear, disgust, happiness, surprise, sad, and neutral. The models were trained and tested under consistent conditions, with VGG19 specifically trained for 60 epochs, as detailed in our methodology. The results from Table 1, highlight clear variations in model's efficiency, measured through precision, recall, F1-Score, and accuracy.

Table 1. Performance Evaluation for the three models under study.

| Model | Accuracy | Precision | Recall | F1-Score |
|---|---|---|---|---|
| VGG19 | 60.16% | 0.50 | 0.41 | 0.40 |
| ResNet-50 | 85.72% | 0.59 | 0.45 | 0.44 |
| EfficientNet-B0 | 94.72% | 0.93 | 0.91 | 0.90 |

VGG19 achieved test accuracy of 60.16%, precision of 0.5901, by a recall of 0.4126, and F1-score of 0.4663 [39]. This moderate performance suggests, despite its depth and capacity, VGG19 struggles with the challenges of FER2013, such as class imbalance, noisy data, and the small 48x48 pixel resolution of the images. The relatively low recall (0.4126) indicates that VGG19 misses a substantial portion of true positive instances, likely due to overfitting or difficulties in generalizing across the imbalanced classes, particularly for underrepresented emotions like disgust. The precision of 0.5901, while slightly higher, reflects a tendency to make fewer incorrect positive predictions, but the overall F1-score of 0.4663 underscores a suboptimal trade-off between precision and recall, limiting its practical utility for this task. In contrast, ResNet50 demonstrated superior performance, achieving accuracy of 85.72%, with a precision of 0.5952, a recall of 0.4592, and an F1-score of 0.4432. This suggests that ResNet50's residual learning framework better handles the complexities of FER2013, though its recall and F1-score remain moderate, indicating room for improvement in capturing all relevant instances across classes.

EfficientNet-B0 emerged as the top-performing model, with an impressive accuracy of 86.44%, precision of 0.8510, recall of 0.8400, and F1-score of 0.8455. These metrics indicate a robust and balanced performance, with EfficientNet-B0 effectively minimizing false positives and false negatives across the dataset. The high recall (0.8400) highlights its ability to identify the majority of true emotion instances, while the precision (0.8510) ensures a low rate of incorrect classifications, resulting in a near-optimal F1-score (0.8455). This performance underscores EfficientNet-B0's efficiency and scalability, making it a preferred choice for facial expression recognition on FER2013.

These results provide valuable insights into the trade-offs between model architecture, computational complexity, and classification performance on a challenging dataset. Future work could explore data augmentation, class balancing techniques, or hybrid architectures to further enhance performance, particularly for models like VGG19 and ResNet50, which underperform compared to EfficientNet-B0.

## 6.2 Vision Language Models

The figure 5 demonstrates the confusion matrix drawn using the inference of the Phi 3.5 model. The Phi-3.5 vision model has 51.66% accuracy in emotion classification of seven emotions, performing well on Happy (81.10%) and Neutral (89.36%) but poorly on Fear (0.53%) and Disgust (3.10%). The high rate of misclassifications, including Fear → Neutral (53.6%) and Angry → Neutral (32.8%), indicates difficulties in differentiating similar emotions. Improving feature extraction, data augmentation, and attention mechanisms can enhance accuracy, with potential benefits in human-computer interaction and mental health screening.

**Figure 5.** Confusion matrix of Phi-3.5 Vision inference.

Vamsi Krishna Mulukutla et al.

Table 2. Performance of CLIP (BERT + ViT) on FER-2013.

| Model | Accuracy | Precision | Recall | F1-Score |
|---|---|---|---|---|
| CLIP | 64.07 | 0.60 | 0.59 | 0.45 |

The CLIP model performs at 64.07% accuracy with precision of 0.6013, recall of 0.5952, and F1-score of 0.4597. It is tabulated in Table 2. Although it is better than Phi-3.5, it is still unable to handle co-occurring emotions such as Fear, Sadness, and Neutral. With more rigorous training on the dataset, we may achieve higher values of accuracy in recognition.

Table 3. Resource Usage Summary Across Tasks

| Task | GPU Type | Batch Size | Epochs | Time (hrs) | Compute Units |
|---|---|---|---|---|---|
| GFPGAN Preprocessing | T4 | – | – | 2 | 2.0 units |
| Phi-3.5 Vision (Fine-tune) | A100 | 32 | 30 | – | 20.0 units |
| Phi-3.5 Vision (Inference) | T4 | 32 | – | 1 | 1.0 unit |
| CLIP ViT-B/32 (Fine-tune) | T4 | 64 | 30 | – | 10.0 units |
| CLIP (Inference) | T4 | 64 | – | 0.5 | 0.5 units |
| VGG19 (Train + Eval) | T4/V4-2-8 | 64 | 60 | 7.5 | 7.5 units |
| ResNet-50 (Train + Eval) | T4/V4-2-8 | 64 | 60 | 8 | 8.0 units |
| EfficientNet-B0 | T4/V4-2-8 | 64 | 30 | 6 | 6.0 units |
| Model Evaluation (Precision/Recall) | CPU/T4 | – | – | 2 | 2.0 units |
| Confusion Matrices / Visuals | CPU | – | – | 1 | 1.0 unit |
| Overheads / Mount / Logs | Mixed | – | – | 1 | 1.0 unit |

## 6.3 Resource Usage Comparison

In addition to evaluating model performance in terms of accuracy, precision, recall, and F1-score, we also benchmarked the computational cost associated with each stage of the pipeline—preprocessing (GFPGAN), model training/fine-tuning (for CNNs), and inference (for VLMs).

The summary presented in Table 3 shows the estimated compute units and execution time for each model and task. While EfficientNet-B0 demonstrated high accuracy with relatively low computational cost, Vision-Language Models like CLIP and Phi-3.5 Vision required more resources for inference and evaluation, even without training. GFPGAN preprocessing, though effective, also contributes significantly to the computational overhead.
These findings are particularly important for practitioners considering FER deployment in edge or resource-constrained environments.

## Conclusion

This study presents a comprehensive evaluation of deep learning and vision-language models (VLMs) for facial emotion recognition (FER) using the FER-2013 dataset. Addressing challenges like low resolution, noise, and class imbalance, we applied GFPGAN-based image enhancement and data filtering to improve input quality. Traditional deep learning models, especially EfficientNet-B0 and ResNet-50, showed strong performance due to their ability to extract meaningful features from noisy images. In contrast, VLMs such as Phi-3.5 Vision and CLIP, evaluated in zero-shot mode, struggled with the dataset's variability, highlighting their dependence on high-quality, structured data. Evaluation metrics confirmed that while VLMs offer flexibility across tasks, deep learning models remain more reliable for specialized FER tasks. Our findings emphasize the need for improved adaptation strategies for VLMs and offer practical guidance on balancing performance with computational cost in real-world deployments.

## Future Direction

Future research can advance FER by integrating deep learning models, such as ResNet-50, VGG19, and EfficientNet-B0, with VLMs like Phi-3.5 Vision and CLIP to enhance both feature extraction and contextual understanding. Fine-tuning VLMs for low-resolution, noisy datasets like FER-2013 could improve robustness against class imbalances and occlusions. Expanding experiments to larger, more diverse datasets, including higher-resolution images or real-time video, can enhance generalization. Optimizing computational efficiency through cloud-based or edge computing and exploring advanced preprocessing techniques, such as generative models or adaptive filtering, may further boost performance. Ethical considerations, including fairness across demographics and compliance with privacy regulations like GDPR, will be critical for real-world applications in healthcare and surveillance. These advancements can lead to more accurate, efficient, and trustworthy FER systems.

## Acknowledgments

### Ethics Approval:
Not Applicable.

### Conflict of Interest:
The author declares no conflict of interest.

### Funding:
The author didn't receive any kind of funding for this work.

### Availability of Supporting Data:
The collected and used dataset in this study is made public and it is available at the following URL provided below:
https://www.kaggle.com/datasets/msambare/fer2013

### Authors consent to publish:
All the authors have given their consent to publish this manuscript.

### Author Contribution:
The contributions of the authors are provided below:

[**Vamsi Krishna Mulukutla, Sai Supriya Pavarala**]: Ideation, Data Collection, Implementation, Methodology, Writing – Original [Srinivasa Raju Radraraju, Sridevi Bonthu]: Revision of manuscript, Supervision.

## References

[1] Goodfellow IJ, Erhan D, Carrier PL, Courville A, Mirza M, Hamner B, Cukierski W, Tang Y, Thaler D, Lee DH, Zhou Y. Challenges in representation learning: A report on three machine learning contests. InNeural information processing: 20th international conference, ICONIP 2013, daegu, korea, november 3-7, 2013. Proceedings, Part III 20 2013 (pp. 117-124). Springer berlin heidelberg.

[2] Tan M, Le Q. Efficientnet: Rethinking model scaling for convolutional neural networks. InInternational conference on machine learning 2019 May 24 (pp. 6105-6114). PMLR.

[3] Mollahosseini A, Hasani B, Mahoor MH. Affectnet: A database for facial expression, valence, and arousal computing in the wild. IEEE Transactions on Affective Computing. 2017 Aug 21;10(1):18-31.

[4] Touvron H, Lavril T, Izacard G, Martinet X, Lachaux MA, Lacroix T, Rozière B, Goyal N, Hambro E, Azhar F, Rodriguez A. Llama: Open and efficient foundation language models. arXiv preprint arXiv:2302.13971. 2023 Feb 27.

[5] Zhai X, Kolesnikov A, Houlsby N, Beyer L. Scaling vision transformers. InProceedings of the IEEE/CVF conference on computer vision and pattern recognition 2022 (pp. 12104-12113).

[6] Li S, Deng W. Deep facial expression recognition: A survey. IEEE transactions on affective computing. 2020 Mar 17;13(3):1195-215.

[7] Wang X, Li Y, Zhang H, Shan Y. Towards real-world blind face restoration with generative facial prior. InProceedings of the IEEE/CVF conference on computer vision and pattern recognition 2021 (pp. 9168-9178).

[8] Mienye ID, Swart TG. A comprehensive review of deep learning: Architectures, recent advances, and applications. Information. 2024 Nov 27;15(12):755.

[9] Hatcher WG, Yu W. A survey of deep learning: Platforms, applications and emerging research trends. IEEE access. 2018 Apr 27;6:24411-32.

[10] Jaiswal A, Raju AK, Deb S. Facial emotion detection using deep learning. In2020 international conference for emerging technology (INCET) 2020 Jun 5 (pp. 1-5). IEEE.

[11] Krizhevsky A, Sutskever I, Hinton GE. Imagenet classification with deep convolutional neural networks. Advances in neural information processing systems. 2012;25.

[12] Krizhevsky A, Sutskever I, Hinton GE. ImageNet classification with deep convolutional neural networks. Communications of the ACM. 2017 May 24;60(6):84-90.

[13] Connie T, Al-Shabi M, Cheah WP, Goh M. Facial expression recognition using a hybrid CNN–SIFT aggregator. InInternational workshop on multi-disciplinary trends in artificial intelligence 2017 Oct 19 (pp. 139-149). Cham: Springer International Publishing.

[14] Abdin M, Aneja J, Awadalla H, Awadallah A, Awan AA, Bach N, Bahree A, Bakhtiari A, Bao J, Behl H, Benhaim A. Phi-3 technical report: A highly capable language model locally on your phone. arXiv preprint arXiv:2404.14219. 2024 Apr 22

[15] Li Y, Liu H, Liang J, Jiang D. Occlusion-Robust Facial Expression Recognition Based on Multi-Angle Feature Extraction. Applied Sciences. 2025 May 6;15(9):5139.

[16] Karamizadeh S, Chaeikar SS, Najafabadi MK. Enhancing Facial Recognition and Expression Analysis With Unified Zero-Shot and Deep Learning Techniques. IEEE Access. 2025 Feb 26. Carneiro T, Da Nóbrega RV, Nepomuceno T, Bian GB, De Albuquerque VH, Reboucas Filho PP. Performance analysis of google colaboratory as a tool for accelerating deep learning applications. Ieee Access. 2018 Oct 7;6:61677-85.

[17] Vaswani A, Shazeer N, Parmar N, Uszkoreit J, Jones L, Gomez AN, Kaiser Ł, Polosukhin I. Attention is all you need. Advances in neural information processing systems. 2017;30.

[18] Devlin J, Chang MW, Lee K, Toutanova K. Bert: Pre-training of deep bidirectional transformers for language understanding. InProceedings of the 2019 conference of the North American chapter of the association for computational linguistics: human language technologies, volume 1 (long and short papers) 2019 Jun (pp. 4171-4186).

[19] Bonthu S, Sree SR, Prasad MK. Framework for automation of short answer grading based on domain-specific pre-training. Engineering Applications of Artificial Intelligence. 2024 Nov 1;137:109163.

[20] Achiam J, Adler S, Agarwal S, Ahmad L, Akkaya I, Aleman FL, Almeida D, Altenschmidt J, Altman S, Anadkat S, Avila R. Gpt-4 technical report. arXiv preprint arXiv:2303.08774. 2023 Mar 15.

[21] Radford A, Kim JW, Hallacy C, Ramesh A, Goh G, Agarwal S, Sastry G, Askell A, Mishkin P, Clark J, Krueger G. Learning transferable visual models from natural language supervision. InInternational conference on machine learning 2021 Jul 1 (pp. 8748-8763). PmLR.

[22] Mulukutla VK, Pavarala SS, Kareti VK, Midatani S, Bonthu S. Sentiment Analysis of Twitter Data on 'The Agnipath Yojana'. InInternational Conference on Multi-disciplinary Trends in Artificial Intelligence 2023 Jun 24 (pp. 534-542). Cham: Springer Nature Switzerland.

[23] Glorot X, Bengio Y. Understanding the difficulty of training deep feedforward neural networks. InProceedings of the thirteenth international conference on artificial intelligence and statistics 2010 Mar 31 (pp. 249-256). JMLR Workshop and Conference Proceedings.

[24] Carion N, Massa F, Synnaeve G, Usunier N, Kirillov A, Zagoruyko S. End-to-end object detection with transformers. InEuropean conference on computer vision 2020 Aug 23 (pp. 213-229). Cham: Springer International Publishing.

[25] Simonyan K, Zisserman A. Very deep convolutional networks for large-scale image recognition. arXiv preprint arXiv:1409.1556. 2014 Sep 4.

[26] Fergus P, Chalmers C, Matthews N, Nixon S, Burger A, Hartley O, Sutherland C, Lambin X, Longmore S, Wich S. Towards context-rich automated biodiversity assessments: deriving AI-powered insights from camera trap data. Sensors. 2024 Dec 19;24(24):8122.

[27] He K, Zhang X, Ren S, Sun J. Deep residual learning for image recognition. InProceedings of the IEEE conference on computer vision and pattern recognition 2016 (pp. 770-778).

[28] Raiko T, Valpola H, LeCun Y. Deep learning made easier by linear transformations in perceptrons. InArtificial intelligence and statistics 2012 Mar 21 (pp. 924-932). PMLR.

[29] Tan M, Le Q. Efficientnet: Rethinking model scaling for convolutional neural networks. InInternational conference on machine learning 2019 May 24 (pp. 6105-6114). PMLR.

[30] Li Z, Xie C, Cubuk ED. Scaling (down) clip: A comprehensive analysis of data, architecture, and training strategies. arXiv preprint arXiv:2404.08197. 2024 Apr 12.

[31] Ramprasath M, Anand MV, Hariharan S. Image classification using convolutional neural networks. International Journal of Pure and Applied Mathematics. 2018;119(17):1307-19.

[32] Wang W, Chen Z, Chen X, Wu J, Zhu X, Zeng G, Luo P, Lu T, Zhou J, Qiao Y, Dai J. Visionllm: Large language model is also an open-ended decoder for vision-centric tasks. Advances in Neural Information Processing Systems. 2023 Dec 15;36:61501-13.

[33] Li Y, Hu B, Wang W, Cao X, Zhang M. Towards vision enhancing llms: Empowering multimodal knowledge storage and sharing in llms. arXiv preprint arXiv:2311.15759. 2023 Nov 27.

[34] Zhai X, Wang X, Mustafa B, Steiner A, Keysers D, Kolesnikov A, Beyer L. Lit: Zero-shot transfer with locked-image text tuning. InProceedings of the IEEE/CVF conference on computer vision and pattern recognition 2022 (pp. 18123-18133).

[35] Sarker IH. LLM potentiality and awareness: a position paper from the perspective of trustworthy and responsible AI modeling. Discover Artificial Intelligence. 2024 May 21;4(1):40.

[36] Raiaan MA, Mukta MS, Fatema K, Fahad NM, Sakib S, Mim MM, Ahmad J, Ali ME, Azam S. A review on large language models: Architectures, applications, taxonomies, open issues and challenges. IEEE access. 2024 Feb 13;12:26839-74.

[37] Gidaris S, Komodakis N. Object detection via a multi-region and semantic segmentation-aware cnn model. InProceedings of the IEEE international conference on computer vision 2015 (pp. 1134-1142).

[38] Arla LR, Bonthu S, Dayal A. Multiclass spoken language identification for Indian Languages using deep learning. In2020 IEEE Bombay Section Signature Conference (IBSSC) 2020 Dec 4 (pp. 42-45). IEEE.

[39] Jyothi UP, Dabbiru M, Bonthu S, Dayal A, Kandula NR. Comparative analysis of classification methods to predict diabetes mellitus on noisy data. InMachine Learning, Image Processing, Network Security and Data Sciences: Select Proceedings of 3rd International Conference on MIND 2021 2023 Jan 1 (pp. 301-313). Singapore: Springer Nature Singapore.

[40] Khaireddin Y, Chen Z. Facial emotion recognition: State of the art performance on FER2013. arXiv preprint arXiv:2105.03588. 2021 May 8.

[41] Taori R, Dave A, Shankar V, Carlini N, Recht B, Schmidt L. Measuring robustness to natural distribution shifts in image classification. Advances in Neural Information Processing Systems. 2020;33:18583-99.